%% file: iclr2027_paper.tex
\documentclass{article}

\usepackage{iclr2027_conference,times}
\input{math_commands.tex}

\usepackage{amsmath,amssymb,mathtools}
\usepackage{graphicx}
\usepackage{booktabs}
\usepackage{multirow}
\usepackage{enumitem}
\usepackage{xcolor}
\usepackage{microtype}
\usepackage{placeins}
\usepackage{float}
\usepackage{hyperref}
\usepackage{url}
\hypersetup{hidelinks}
\graphicspath{{figures/}}

\title{From Association to Decision Value:\\
Testing Paired Instruction Dependence for VLA Corpus Selection}

\author{Nurmukhambet Serikbayev, Abdul Basit, Muhammad Shafique \\
\{nms9668, abdul.basit,  muhammad.shafique\}@nyu.edu\\
\textit{eBRAIN Lab, Division of Engineering, NYUAD}, Abu Dhabi, UAE
}
\iclrfinalcopy

\begin{document}

\maketitle

\input{Sections/01_abstract}

\begin{figure}[H]
  \centering
  \includegraphics[width=\linewidth]{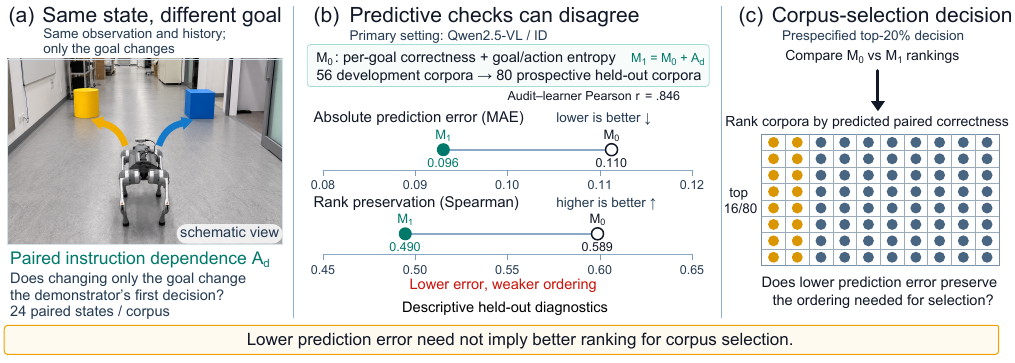}
  \caption{\textbf{Why prediction is not yet selection;} (a) A schematic third-person view illustrates the same-observation goal swap used to test whether the demonstrator changes its first response. (b) In the primary held-out test, adding $A_d$ lowers MAE $.110\!\to\!.096$ while Spearman falls $.589\!\to\!.490$. (c) The prespecified top-20\% rule ranks 80 corpora and selects 16, so ordering determines utility.}
  \label{fig:motivation}
\end{figure}
\input{Sections/02_introduction}
\input{Sections/03_related_works}
\input{Sections/04_methodology}
\input{Sections/05_experimental_setup}
\input{Sections/06_results}
\input{Sections/07_conclusion}

\section*{Reproducibility Statement}
The supplement specifies corpus construction, prompts, seeds, partitions, runtime, and statistical decisions. The artifact provides code, ordered split identifiers, predictions.

\section*{Ethics Statement}
The corpus study and 768-episode closed-loop evaluation use simulation. The 48 physical Go2 runs used operator supervision and a depth-based safety mask; no in-run intervention was logged. These descriptive trials do not establish safety beyond the recorded environments. Deployment requires oversight, emergency stopping, and site-specific validation.

\section*{AI Use Statement}
Language models assisted with code refinement and grammar review. The authors verified all suggested changes and remain fully responsible for the manuscript content.

\bibliography{cite}
\bibliographystyle{iclr2027_conference}

\appendix
\input{Sections/08_appendix}

\end{document}

%% file: math_commands.tex
\usepackage{amsmath,amsfonts,bm}

\def\eqref#1{equation~\ref{#1}}

\def\1{\bm{1}}

\DeclareMathAlphabet{\mathsfit}{\encodingdefault}{\sfdefault}{m}{sl}
\SetMathAlphabet{\mathsfit}{bold}{\encodingdefault}{\sfdefault}{bx}{n}



%% file: Sections/01_abstract.tex
\begin{abstract}
Choosing among robot demonstration corpora is a ranking problem: an upstream diagnostic must preserve which candidates produce better downstream policies, not merely correlate with average performance. We study \emph{paired instruction dependence}, the rate at which a demonstrator changes its first decision when only the goal changes. Across 136 independently seeded quadruped-navigation corpora, predictions developed on 56 corpora are evaluated prospectively on 80 held-out corpora. The audit is reliable and associated with downstream paired correctness, and lowers mean absolute prediction error when added to a correctness-aware baseline, but the primary proper-log-score increment is $+0.059$ nats (95\% corpus-bootstrap CI $[-0.027,0.133]$) and the prespecified top-20\% comparison does not establish an incremental selection gain. An exploratory response decomposition suggests a structural source of this predictive--ranking mismatch: under the tested predictor specification, separating both-correct and both-wrong responses yields larger held-out gains than their unsigned switch statistic. Seed-to-seed agreement indicates that the held-out endpoint retains recoverable rank signal. These results show that counterfactual response structure can be reliable and predictive while providing limited incremental decision value in this setting when compressed into an unsigned statistic, motivating diagnostics that preserve outcome-relevant structure and are validated against the decisions they are intended to support.
\end{abstract}

%% file: Sections/02_introduction.tex
\section{Introduction}
\label{sec:introduction}

The downstream value of a robot demonstration corpus is usually known only after a vision--language--action (VLA) policy has been fine-tuned on it. Evaluating every candidate therefore pays much of the cost that an upstream diagnostic is meant to avoid. The practical question is consequently decision-focused: can a measurement rank candidate corpora well enough to select the ones that will produce better policies?

Language-conditioned behavior creates a tempting diagnostic. If two goals require opposite turns from the same observation, a demonstrator that follows language should usually change its decision when the goal changes. We call the resulting switch rate \emph{paired instruction dependence}, $A_d$. Unlike ordinary correctness, it uses a controlled same-state intervention and tests whether the response depends on the instruction. This distinction is relevant to generalist VLAs, whose success can otherwise conceal shortcut policies that weakly use language~\citep{geirhos2020shortcut,brohan2023rt2,kim2024openvla,octo2024,black2024pi0}.

Dependence, however, is unsigned. A demonstrator can switch in the correct direction for both goals or in the wrong direction. With binary left/right responses, $A_d$ is exactly the sum of these both-correct and both-wrong cases; the observed \textsc{halt} action makes that identity approximate rather than exact in our data. The relevant question is therefore narrower than whether language sensitivity matters in general: does unsigned same-state response structure improve prediction and selection after marginal demonstrator correctness is known?

This question also separates four criteria that are often conflated:
\begin{equation*}
\text{reliable measurement}\;\longrightarrow\;\text{association}\;\longrightarrow\;
\text{incremental prediction}\;\longrightarrow\;\text{decision value}.
\end{equation*}
A signal may track broad outcome differences or lower mean error while failing to preserve the ordering near a selection cutoff. Figure~\ref{fig:motivation} illustrates this \emph{decision-value gap}.

We test it prospectively on 136 independently seeded quadruped-navigation corpora. Nested probabilistic predictors are developed on 56 corpora, while 80 disjoint corpora form the held-out test population. The correctness-aware baseline $M_0$ uses per-goal demonstrator correctness and corpus entropies; $M_1$ adds only $A_d$. We train 544 adapters across Qwen2.5-VL-3B and Qwen3.5-VL-0.8B and evaluate in-distribution and under a controlled scene shift. The primary selection endpoint is paired first-decision correctness; closed-loop evaluation separately tests how this local endpoint relates to full navigation. Predictors, held-out corpus identifiers, and the top-20\% operating point were prespecified before held-out outcomes were revealed.

The study makes three main contributions:
\begin{enumerate}[leftmargin=*,nosep]
  \item \textbf{A prospective decision-value evaluation:} we separate reliability, association, proper probabilistic prediction, and top-$k$ selection for a same-state robot-data diagnostic, using the corpus as the unit of inference.
  \item \textbf{Correctness-resolved response structure locates a candidate source of the predictive--ranking mismatch:} the unsigned switch statistic mixes both-correct and both-wrong behavior; separating them yields larger exploratory held-out gains under the tested predictor specification.
  \item \textbf{Decision resolution and practical alternatives:} a cross-seed resolution benchmark establishes recoverable corpus ordering, while demonstrator and proxy baselines locate the audit relative to realistic selectors.
\end{enumerate}

%% file: Sections/03_related_works.tex
\section{Related Work}
\label{sec:related}

\paragraph{Vision--language--action learning:}
Language-conditioned robot learning spans task-specific imitation~\citep{lynch2020language,mees2022hulc,nair2022crowdsourcing}, large pretrained embodied models~\citep{driess2023palme,brohan2023rt2,kim2024openvla,octo2024,black2024pi0}, and heterogeneous robot corpora~\citep{openx2023}. Physical VLM-to-VLA transfer is increasingly practical~\citep{hancock2026actions}; NaVILA separates language-level navigation decisions from low-level legged control~\citep{jiang2024navila}, as we do. We study the preceding data decision: which fixed demonstration corpus should train a policy?

\paragraph{Language grounding and counterfactual evaluation:}
Vision-and-language navigation established grounded instruction following~\citep{anderson2018vision}, but success can hide instruction-insensitive shortcuts. Controlled tests expose memorized action sequences and semantic failures~\citep{fei2025liberoplus,zhou2026liberopro,hou2026langgap,yu2026robosemanticbench}, while mismatched instructions can supply corrective supervision~\citep{kondoh2026counteralign}. Our intervention similarly holds the state fixed while changing the goal, but it measures a data-generating demonstrator and predicts a separately trained learner. Crucially, we distinguish unsigned response change from correctness-resolved response structure.

\paragraph{Robot-data selection and valuation:}
Robot policies are sensitive to demonstration composition and scale~\citep{mandlekar2022what,saxena2025whatmatters,lin2025scaling}. Data valuation uses marginal contribution, training dynamics, gradients, or standardized retraining~\citep{ghorbani2019datashapley,swayamdipta2020dataset,xia2024less,gadre2023datacomp}. DemInf scores demonstrations through state--action mutual information~\citep{hejna2025deminf}; DataMIL learns policy-aware datamodels~\citep{dass2026datamil}. Unlike end-to-end selectors such as DataMIL, we test whether an interpretable pre-training corpus diagnostic retains incremental value at the downstream selection decision. Our unit is an entire candidate corpus, and the nested comparison conditions on marginal demonstrator correctness.

\paragraph{Proxy selection and decision-aware evaluation:}
Selection via Proxy uses a smaller model to rank data efficiently~\citep{coleman2020selection}; compute-aware selection makes the appropriate proxy depend on its cost~\citep{yin2025compute}. Small training runs can preserve data-recipe rankings inconsistently across proxy configurations~\citep{wang2026smallruns}. Upstream probes can correlate with control~\citep{dong2026visualproxy}, and proxy environments can be judged by absolute prediction and rank preservation~\citep{quevedo2026worldgym}. Prior work therefore distinguishes absolute prediction quality from rank preservation and selection utility. We extend this distinction to prospective robot-corpus selection with prespecified decisions, a proper-score test, a seed-based resolution benchmark, and direct comparison with practical selectors.

%% file: Sections/04_methodology.tex
\section{Method}
\label{sec:method}

\begin{figure}[t]
  \centering
  \makebox[\linewidth][c]{\includegraphics[width=1\linewidth]{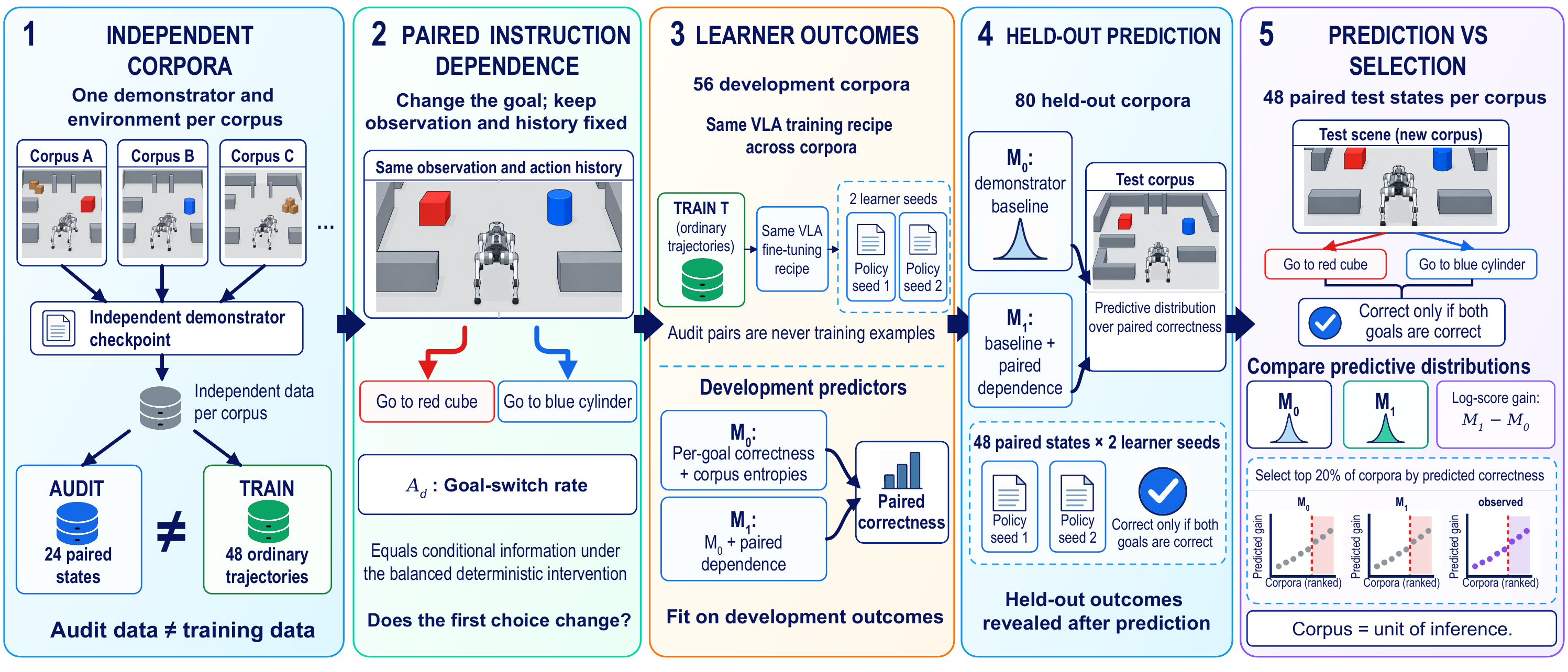}}
  \caption{\textbf{Prospective test of paired response structure;} Audit states and training trajectories are disjoint, and only the goal changes within an audit pair. Baseline $M_0$ uses marginal per-goal correctness and corpus entropies; $M_1$ adds the same-state switch rate $A_d$. Both models predict paired learner correctness and rank the same held-out corpora.}
  \label{fig:method}
\end{figure}

Figure~\ref{fig:method} summarizes the prospective comparison and the disjoint roles of audit, training, and held-out evaluation data.

\subsection{Same-state goal-swap audit}

For each audit state, we query the demonstrator twice with an identical visual observation and action history. Only the object goal changes, and the two goals require opposite branches. Let $d$ index a corpus, $u$ a state, $x^-,x^+$ the goals, and $B_d(u,x)$ the first response. For a deterministic demonstrator, paired instruction dependence is
\begin{equation}
 A_d = \frac{1}{n_A}\sum_{u=1}^{n_A}
 \mathbf{1}\!\left[B_d(u,x^-)\neq B_d(u,x^+)\right].
 \label{eq:audit}
\end{equation}
It is an unsigned response statistic: changing correctly for both goals and changing incorrectly for both goals can each increase $A_d$. Under the balanced deterministic intervention, each unequal pair identifies the goal and contributes one bit, so Eq.~\ref{eq:audit} equals state-averaged conditional mutual information (Appendix~\ref{app:information}). This equivalence does not make $A_d$ a correctness or downstream-value measure.

The audit requires counterfactual access to a queryable demonstrator and is unavailable from a fixed corpus containing only one instruction per state. It is therefore applicable in settings such as automated simulation or online data generation, where the same state can be replayed under alternate goals before committing to a large collection. The study therefore concerns this access model rather than arbitrary human demonstrations.

\subsection{What does the switch rate contain?}

Let $a_d$, $b_d$, and $c_d$ be the fractions of audit states for which both goal responses are correct, both are wrong, or exactly one is correct. Marginal per-goal demonstrator correctness is
\begin{equation}
 p_d=a_d+\tfrac12c_d=\tfrac12+\tfrac12(a_d-b_d).
 \label{eq:abc}
\end{equation}
If responses are restricted to left/right, opposite required branches imply $A_d=a_d+b_d$. Thus, in the binary case, $(p_d,A_d)$ is an invertible reparameterization of $(a_d,b_d)$: $p_d$ contains their signed difference, whereas $A_d$ contains their unsigned sum. Our demonstrators may also emit \textsc{halt}; this breaks the exact $A_d=a_d+b_d$ identity in our data, although Eq.~\ref{eq:abc} remains exact. The decomposition makes the primary question precise: does unsigned joint response structure improve prediction beyond marginal correctness and corpus composition?

\subsection{Nested prediction test}

After fine-tuning with learner seed $s$, let $C_{d,s}\in\{0,\ldots,m\}$ count evaluation states where the VLA predicts the correct branch for \emph{both} goals. This conjunction rejects constant-side policies. We compare nested beta--binomial predictors:
\begin{align}
C_{d,s}&\sim\mathrm{BetaBinomial}(m,\mu_{d,s},\rho), \label{eq:bb}\\
M_0:\quad\mathrm{logit}(\mu_{d,s})&=\alpha+\beta^\top W_d+\gamma_s,\label{eq:m0}\\
M_1:\quad\mathrm{logit}(\mu_{d,s})&=\alpha+\beta^\top W_d+\eta A_d+\gamma_s.\label{eq:m1}
\end{align}
$W_d$ contains marginal per-goal demonstrator correctness, training-goal entropy, and coarse training-action entropy. Both predictors therefore use the same 48 demonstrator responses; $M_1-M_0$ isolates pairing structure rather than extra queries. The crossed seed effect $\gamma_s$ is a nuisance term and $\rho$ captures extra-binomial variation.

Feature scaling and the ridge penalty are learned on development corpora. The primary $M_1$ shares $M_0$'s penalty so that adding $A_d$ does not also add a hyperparameter search. Test-audit uncertainty is marginalized by resampling the 24 paired states. Exploratory checks separately tune the ridge penalty for each response parameterization, decompose $A_d$ into $a_d$ and $b_d$, and add existing rollout features. None changes the prespecified endpoint or primary result.

\subsection{Prediction, ranking, and selection}

The primary estimand is the equal-corpus improvement in proper log score,
\begin{equation}
  \bar\delta=\frac{1}{|\mathcal B|}\sum_{d\in\mathcal B}
  \frac{1}{|\mathcal S|}\sum_{s\in\mathcal S}
  \left[\log p_{1,d,s}(C_{d,s})-\log p_{0,d,s}(C_{d,s})\right],
  \label{eq:delta}
\end{equation}
where $\mathcal B$ is the held-out corpus population and $\mathcal S$ the two learner seeds. A corpus-cluster bootstrap supplies the 95\% interval. The prespecified primary rule requires its lower bound to exceed zero. The $0.105\approx-\log(0.9)$-nat magnitude reference corresponds to multiplying the geometric-mean assigned probability by approximately 1.11; it is not a universal utility threshold. Log score~\citep{gneiting2007strictly} evaluates the full predictive distribution and penalizes confident errors.

For selection, model $M_j$ ranks each corpus by predicted mean correctness,
\begin{equation}
 \widehat q_j(d)=\frac{1}{|\mathcal S|}\sum_{s\in\mathcal S}
 \mathbb E_{M_j}[C_{d,s}/m],
 \label{eq:rankscore}
\end{equation}
and selects $K=16$ of 80 corpora. We report selected correctness, regret relative to the realized oracle set, and oracle overlap. The operating point was prespecified before outcome reveal; its confidence interval is exploratory. No held-out outcome enters fitting, tuning, or ranking.

\subsection{Resolution and comparator analyses}

Finite evaluation states and learner seeds limit any ranking method. We use cross-seed ranking as an empirical resolution benchmark and compare against demonstrator and proxy selectors. We also repeat $M_0/M_1$ comparisons after excluding the two least mature checkpoints. These exploratory analyses do not alter the prespecified primary test.

%% file: Sections/05_experimental_setup.tex
\section{Experimental Setup}
\label{sec:setup}

\subsection{Corpus population}

We study Unitree Go2 object-goal navigation in Isaac Lab~\citep{mittal2023orbit}. Given forward RGB, instruction text, and action history, the VLA emits a discrete navigation command for a fixed low-level controller; only that controller sees proprioception. Every corpus uses a separately trained tabular Q-learning demonstrator with unique training, layout, physics, and rendering seeds. Checkpoints span 10, 50, 250, 1,000, and 50,000 episodes. The held-out population has 16 corpora per level; development has 12 at 10 episodes and 11 at each remaining level. Checkpoint maturity is excluded from the deployable baseline because it is an experimental assignment, not a property available for an arbitrary corpus.

Each corpus contains 24 paired audit states, 48 unpaired multi-step training trajectories, and 48 paired in-distribution (ID) evaluation states. Audit, training, and evaluation packets are disjoint at the state, image, and trajectory levels. All 6,528 attempted training rollouts passed prespecified structural-validity checks, so acceptance is constant across corpora and cannot explain their ranking; first-branch correctness instead ranges from $.375$ to $.813$. Appendix~\ref{app:reproducibility} gives the complete validity checklist.

\begin{figure}[!t]
  \centering
  \includegraphics[width=\linewidth]{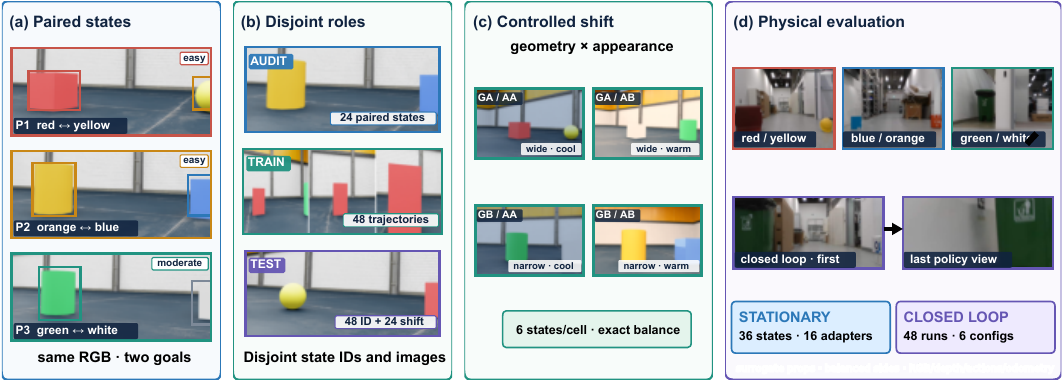}
  \caption{\textbf{Experimental design across simulation and hardware;} (a) Paired states span three semantic families with RGB fixed across goals. (b) Audit, training, and evaluation observations are disjoint. (c) Controlled scene shift balances a $2\!\times\!2$ geometry/appearance design. (d) Stationary real images evaluate 16 adapters; 48 physical runs describe two deployed adapters.}
  \label{fig:setup}
\end{figure}

Across 136 corpora, the study contains 6,528 training trajectories (43,084 steps), 3,264 audit states, and 6,528 ID evaluation states. The 56 development and 80 held-out corpora share no demonstrator seeds or scene realizations. Figure~\ref{fig:setup} summarizes the data roles and evaluations.

The audit requires 48 demonstrator queries and costs a median 4.18 GPU-min per corpus. A Qwen2.5 fit and ID evaluation average 11.46 GPU-min for one seed, or $2.74\times$ the audit cost. Appendix Table~\ref{tab:cost} reports the complete measured cost comparison.

\subsection{VLA training and evaluation}

The primary learner is Qwen2.5-VL-3B-Instruct~\citep{bai2025qwen25vl}; Qwen3.5-VL-0.8B is both a supporting backbone and an exploratory proxy selector~\citep{qwen2026qwen35}. Within each backbone, only corpus and learner seed vary. We hold the vision tower fixed and train rank-8 LoRA adapters~\citep{hu2022lora} on language-model projections. Outputs are left/right turns, forward, halt, or done; invalid generations are errors. The full factorial gives 544 adapters. A Qwen2.5 fit averages 10.68 GPU-min and Qwen3.5 averages 13.41 GPU-min over the measured runs.

\paragraph{ID and controlled shift:}
The ID endpoint has 48 paired states per corpus; both goals must receive the required branch. The 24-state shifted set crosses two spawn/target geometries with two lighting settings, six states per cell. Camera, warehouse, objects, prompt, action space, and controller remain fixed, so this is controlled rather than open-world shift.

\paragraph{Closed-loop simulation:}
For 16 held-out corpora at evenly spaced $A_d$ ranks, the first Qwen2.5 seed runs 24 scenes and both goals (768 episodes). We record first-command correctness, target reach, collision/fall, path efficiency, and recovery. These aggregate rates and exploratory rank correlations describe the stratified subset, not all 80 held-out corpora.

\paragraph{Physical evidence:}
Stationary evaluation uses 36 human-screened Go2 images across three layouts, three surrogate-object pairs, two sides, and two difficulties (72 queries per adapter). Closed-loop deployment uses one adapter per backbone, chosen before physical evaluation by simulation outcomes from the 16 committed roots. The VLA acts from onboard RGB and history; a depth safety mask can veto forward motion. The 48 physical runs are descriptive because terminal state was not externally adjudicated and parts of the recorded protocol departed from the initial design (Appendix~\ref{app:physical}).

\subsection{Inference protocol}

The primary metric is Eq.~\ref{eq:delta}. Secondary diagnostics include absolute error, Pearson/Spearman association, top-20\% correctness, regret, and oracle overlap. Predictors, held-out corpus identifiers, and the top-20\% operating point were prespecified before held-out outcomes were revealed. Inferential intervals resample whole corpora, preserving nested states, paired goals, and crossed seeds. The main endpoint and selection rule were unchanged after reveal; decomposition, resolution, proxy, calibration, and closed-loop correlation analyses are explicitly exploratory. Appendix~\ref{app:reproducibility} records the content-addressed prospective procedure.

%% file: Sections/06_results.tex
\section{Results}
\label{sec:results}

\subsection{Reliability, association, and incremental value diverge}

\begin{figure}[!ht]
  \centering
  \makebox[\linewidth][c]{\includegraphics[width=1.08\linewidth]{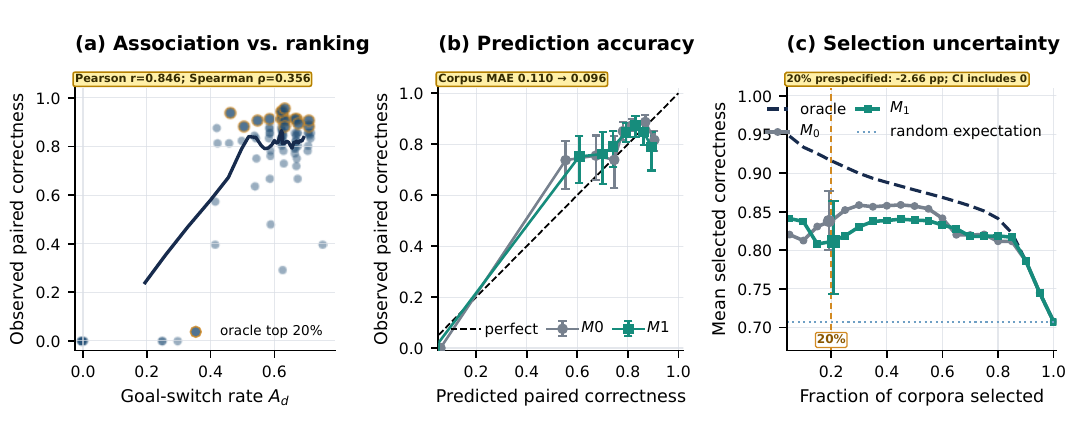}}
  \caption{\textbf{Association, prediction, and selection answer different questions;} (a) Held-out corpora, a 15-corpus moving average, and the realized top 20\%. (b) Equal-count bins with corpus-bootstrap intervals and corpus MAE. (c) The 20\% operating point was prespecified before outcome reveal; its displayed confidence interval and the remaining curve are exploratory.}
  \label{fig:mainresults}
\end{figure}

\begin{figure}[!ht]
  \centering
  \makebox[\linewidth][c]{\includegraphics[width=1.08\linewidth]{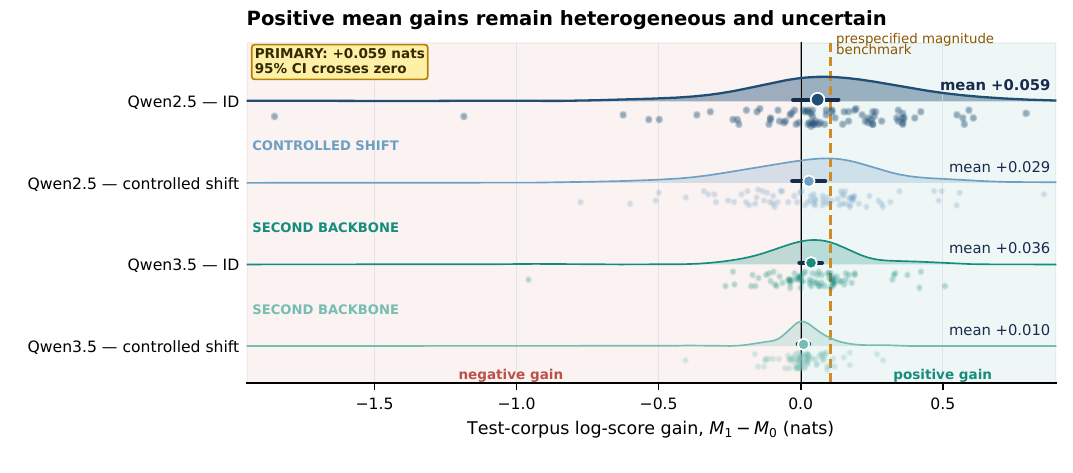}}
  \caption{\textbf{Incremental gains are heterogeneous and uncertain;} Light bands distinguish negative from positive gain. Points are held-out corpora, half-densities show gain distributions, markers are equal-corpus means, and bars are corpus-bootstrap 95\% intervals. Every interval crosses zero and every mean lies below the prespecified magnitude reference.}
  \label{fig:prospective}
\end{figure}

Figures~\ref{fig:mainresults}--\ref{fig:prospective} and Table~\ref{tab:maincrosssetting} summarize the held-out association, predictive, and selection results.

Across all 136 corpora, $A_d$ has Spearman $\rho=.467$ (95\% CI $[.300,.616]$) with Qwen2.5 paired correctness; on the 80 held-out corpora, $\rho=.356$ ($[.111,.565]$). Audit reliability is $.802$ overall and $.766$ on held-out corpora. The audit therefore measures a demonstrator property with substantial estimated reliability and tracks broad learner differences.

The incremental comparison is more uncertain. Adding $A_d$ to $M_0$ gives a Qwen2.5-ID mean log-score gain of $+.059$ nats (95\% corpus-bootstrap CI $[-.027,.133]$). The interval crosses zero, and separately tuning $M_1$'s ridge penalty changes the gain only to $+.046$ ($[-.102,.171]$). The same pattern appears under controlled shift and with Qwen3.5: mean gains are positive but small, and all intervals include zero (Table~\ref{tab:maincrosssetting}). Central predictive intervals are conservative rather than globally overconfident: 80\% coverage is $.925$ for $M_0$ and $.938$ for $M_1$ (Appendix Table~\ref{tab:calibration}).

\begin{table}[H]
\caption{\textbf{Cross-setting summary;} gain is $M_1-M_0$ log score. Arrows show $M_0\!\to\!M_1$; lower MAE and higher Spearman/selected correctness are better.}
\label{tab:maincrosssetting}
\centering
\small
\setlength{\tabcolsep}{3.4pt}
\begin{tabular}{lcccc}
\toprule
Setting & Gain [95\% CI] & MAE & Spearman & Selected \\
\midrule
Qwen2.5 ID    & $+.059\ [-.027,.133]$ & $.110\!\to\!.096$ & $.589\!\to\!.490$ & $.838\!\to\!.811$ \\
Qwen2.5 shift & $+.029\ [-.030,.086]$ & $.106\!\to\!.097$ & $.460\!\to\!.391$ & $.945\!\to\!.917$ \\
Qwen3.5 ID    & $+.036\ [-.005,.075]$ & $.077\!\to\!.073$ & $.752\!\to\!.745$ & $.965\!\to\!.971$ \\
Qwen3.5 shift & $+.010\ [-.010,.030]$ & $.080\!\to\!.078$ & $.729\!\to\!.723$ & $.969\!\to\!.964$ \\
\bottomrule
\end{tabular}
\end{table}

\subsection{Unsigned switching hides correctness-resolved structure}

The demonstrator emits \textsc{halt} in 10.9\% of audit queries, and 107/136 corpora contain at least one nonbinary pair. Thus the binary identity $A_d=a_d+b_d$ is not exact: $A_d-(a_d+b_d)$ averages $.100$ and reaches $.250$. The marginal-correctness identity in Eq.~\ref{eq:abc} is exact. At the 10-episode checkpoint, mean $(a_d,b_d,c_d)=(.156,.016,.827)$; at 50 episodes it is $(.389,.074,.537)$. Both-wrong behavior is present but does not dominate these checkpoint averages.

The decomposition identifies a candidate source of the mismatch. In exploratory held-out tests, $M_0+a_d$ gains $+.104$ nats ($[.049,.153]$) and $M_0+b_d$ gains $+.110$ ($[.022,.188]$), each exceeding the $+.059$ gain from their unsigned, \textsc{halt}-affected mixture. Their selected correctness is $.846$ and $.842$, compared with $.811$ for $M_1$. Under the same exploratory modeling protocol, these correctness-resolved parameterizations yield larger held-out gains than the unsigned switch rate. This ordering persists when the ridge penalty is selected separately on development corpora: the selected penalty is $1$ for all three alternatives, with gains $+.126$ for $a_d$, $+.123$ for $b_d$, and $+.046$ for $A_d$ (Appendix~\ref{app:officialdiagnostics}). The comparison remains post-hoc and can reflect the nonbinary \textsc{halt} cases, finite-sample estimation, or regularized model geometry; it does not establish an information-theoretic advantage.

\subsection{The endpoint contains resolvable rank signal}

The two Qwen2.5 learner seeds have rank agreement $\rho=.567$ across 80 corpora. Selecting 16 corpora by seed~1 and evaluating with seed~2 gives correctness $.874$, oracle overlap $.50$, and regret $.057$; reversing the seeds gives the same selected correctness, $.50$ overlap, and $.052$ regret. This empirical resolution benchmark is imperfect, but it is materially above both $M_0$ ($.838$ selected) and $M_1$ ($.811$), so finite-state and learner-seed variation do not fully explain the selection result.

The ranking gap persists after excluding the 10- and 50-episode checkpoints that create the clearest low-performance cluster. Among the remaining 48 corpora, $M_0$ has Spearman $.349$ and selected correctness $.821$, whereas $M_1$ has Spearman $.172$ and selected correctness $.791$. The ordering problem is therefore not solely an artifact of immature demonstrators.

\subsection{Practical baselines locate the audit's decision value}

Table~\ref{tab:decisionbaselines} compares all methods on the same Qwen2.5 held-out outcomes. Demonstrator closed-loop success measured on existing simulator rollouts is inexpensive but weak ($\rho=.281$). Training-trajectory first-branch correctness, also available from the existing simulator rollouts, improves proper score by $+.156$ nats ($[.078,.225]$), yet does not improve top-20\% selection over $M_0$. Structural rollout acceptance is 100\% for every corpus and adds no information.

Qwen3.5 provides the highest observed rank correlation among the tested selectors: pooled outcomes correlate $\rho=.715$ with Qwen2.5 and select corpora with $.866$ Qwen2.5 correctness. Even one Qwen3.5 seed reaches $\rho=.644$--$.662$ and $.846$--$.874$ selected correctness. In this regime, the proxy therefore preserves the target-model ordering better than either audit-based predictor. It does not establish a cost win: a measured Qwen3.5 fit averages 13.41 GPU-min, versus 4.18 for the audit and 11.46 for one Qwen2.5 fit plus evaluation. At these measured costs, an economic advantage for the audit would require amortizing its queries across multiple downstream learners or collection decisions rather than replacing a single inexpensive adapter fit.

\begin{table}[H]
\caption{\textbf{Exploratory selection baselines on Qwen2.5-ID;} all methods rank the same 80 corpora; $K=16$. Gain is proper log score relative to the prespecified $M_0$ when defined. The seed-to-seed row is an empirical resolution benchmark rather than a deployable selector.}
\label{tab:decisionbaselines}
\centering
\scriptsize
\setlength{\tabcolsep}{2.5pt}
\begin{tabular}{lrrrrr}
\toprule
Selector & Gain [95\% CI] & Spearman & Selected & Regret & Overlap \\
\midrule
$M_0$ & --- & .589 & .838 & .078 & .312 \\
$M_1=M_0+A_d$ & $+.059\ [-.027,.133]$ & .490 & .811 & .105 & .250 \\
$M_0+a_d$ & $+.104\ [.049,.153]$ & .560 & .846 & .070 & .312 \\
$M_0+b_d$ & $+.110\ [.022,.188]$ & .531 & .842 & .074 & .250 \\
$M_0+$ training first-correct & $+.156\ [.078,.225]$ & .584 & .831 & .085 & .312 \\
Demonstrator rollout success & --- & .281 & .838 & .078 & .312 \\
Qwen3.5 outcomes & --- & .715 & .866 & .050 & .438 \\
Qwen2.5 seed 1 $\to$ seed 2 & --- & .567 & .874 & .057 & .500 \\
\bottomrule
\end{tabular}
\end{table}

\subsection{Local response and navigation success diverge}

Across 768 simulated episodes, first-command correctness is $.829$ (637/768), while target reach is $.655$ (503/768), a 17.4-point gap ($[10.4,24.9]$). Of 637 correct-first episodes, 134 fail later; none of 131 wrong-first episodes recovers. Across the 16 corpora, target-reach rank correlations are $.368$ for $A_d$, $.361$ for $M_0$, and $.423$ for $M_1$. These small-sample descriptive values do not establish a corpus selector for embodied success. They show that the local endpoint and full navigation are distinct.

On 48 physical Go2 runs, the first choice is correct in 47, but strict terminal success is not estimable under the recorded protocol. We therefore retain physical results as qualitative transfer evidence and separate objective logs, operator reports, interventions, and protocol departures in Appendix~\ref{app:physical}.

\section{Scope and Limitations}
\label{sec:scope}

The audit requires counterfactual demonstrator access and is tested on one simulator, one tabular demonstrator family, short corpora, and a largely maturity-driven population. The resulting claim is specific to this controlled corpus population. Generalization to different demonstrator families, corpus-generation mechanisms, and larger real-world corpora remains untested.

$A_d$ tests one unsigned additive summary. Correctness-resolved parameterizations yield larger exploratory gains under the tested predictor specification, including after separate development-set penalty tuning, but these alternatives were not prespecified and do not by themselves identify the cause of that difference. Corpus value is defined relative to the fixed learner architecture and fine-tuning recipe used in each comparison. Qwen3.5 is a related backbone rather than an independent architecture family, and the proxy comparison uses outcomes collected for this study. Finally, closed-loop and physical evaluations support endpoint interpretation but do not enlarge the evidence for held-out corpus selection.

%% file: Sections/07_conclusion.tex
\section{Conclusion}
\label{sec:conclusion}

Paired instruction dependence provides a controlled case study of how proxy structure can be lost through compression. The switch rate is measured with substantial reliability and associated with learner behavior, but its unsigned form mixes both-correct and both-wrong responses; correctness-resolved parameterizations yield larger exploratory held-out gains under the tested predictor specification. Cross-seed agreement shows that corpus ordering remains recoverable. Together, these results motivate preserving outcome-relevant structure in counterfactual data diagnostics and validating them against the decisions they are intended to support.

%% file: Sections/08_appendix.tex
\section{Information-Theoretic Motivation}
\label{app:information}

For completeness, let $q(U,X,B)$ be the demonstration distribution and define the best achievable log losses with and without the goal:
\begin{align}
\mathcal R_U^*&=\inf_f\;\mathbb E_q[-\log_2 f(B\mid U)], &
\mathcal R_{U,X}^*&=\inf_g\;\mathbb E_q[-\log_2 g(B\mid U,X)].
\end{align}
Expected log loss conditioned on the inputs equals the entropy of the true conditional distribution plus a nonnegative KL divergence. It is minimized by $f^*(b\mid u)=q(b\mid u)$ and $g^*(b\mid u,x)=q(b\mid u,x)$, giving
\begin{equation}
\mathcal R_U^*-\mathcal R_{U,X}^*
=H_q(B\mid U)-H_q(B\mid U,X)=I_q(B;X\mid U).
\end{equation}
Thus $I_q(B;X\mid U)$ measures the ideal log-loss value of the goal for predicting demonstration actions. Under the balanced deterministic intervention in the main study, each state contributes exactly zero or one bit, so their empirical average is $A_d$. Outside that setting, switch rate and conditional mutual information need not coincide. This is a statement about paired demonstrator responses, not a theorem about optimization, finite-model learning, action correctness, or navigation success.

\section{Endpoint and Predictor Details}
\label{app:predictor_details}

For evaluation state $u$, let $B_u^-$ and $B_u^+$ be the required branches and $\widehat B_{d,s}$ the VLA prediction. The count modeled in Eq.~\ref{eq:bb} is
\begin{equation}
C_{d,s}=\sum_{u=1}^{m}\mathbf{1}\!\left[
\widehat B_{d,s}(u,x^-)=B_u^-\;\land\;
\widehat B_{d,s}(u,x^+)=B_u^+
\right].
\end{equation}
This conjunction rejects constant-side policies that can attain deceptively high per-goal accuracy. It also keeps the local decision endpoint separate from closed-loop target reach.

For the top-$K$ decision, let $\bar Y_d$ be realized correctness for test corpus $d$ averaged over the two learner seeds, and let $q_M(d)$ be its score under predictor $M$. Define the selected and oracle sets as
\begin{align}
\widehat{\mathcal T}_M&=\operatorname{TopK}_{d\in\mathcal B}q_M(d), &
\mathcal T^*&=\operatorname{TopK}_{d\in\mathcal B}\bar Y_d.
\end{align}
The reported selected correctness is $K^{-1}\sum_{d\in\widehat{\mathcal T}_M}\bar Y_d$, regret is the oracle mean minus this quantity, and oracle overlap is $|\widehat{\mathcal T}_M\cap\mathcal T^*|/K$. Score ties are broken deterministically by corpus identifier. These definitions apply only to source-level selection among the 80 fixed test corpora.

We use the mean--intraclass-correlation parameterization
\begin{equation}
 \alpha^{\mathrm{BB}}_{d,s}=\mu_{d,s}\frac{1-\rho}{\rho},\qquad
 \beta^{\mathrm{BB}}_{d,s}=(1-\mu_{d,s})\frac{1-\rho}{\rho},
 \qquad C_{d,s}\sim\operatorname{BetaBinomial}
 \!\left(m,\alpha^{\mathrm{BB}}_{d,s},\beta^{\mathrm{BB}}_{d,s}\right).
 \label{eq:bbparameterization}
\end{equation}
Each fitted predictor has one unpenalized dispersion scalar $\rho$, shared over all development corpus--seed rows within that fit; $M_0$ and $M_1$, backbones, and evaluation settings are fitted separately. We estimate regression coefficients and $\rho$ by penalized maximum likelihood on the 56 development corpora using L-BFGS-B. The ridge penalty applies only to standardized scientific features: the intercept, learner-seed nuisance effect, and $\rho$ are unpenalized.

The varying columns of $W_d$ are marginal per-goal demonstrator correctness, instruction entropy, and action entropy. The ridge value $\lambda=10$ is selected by whole-corpus cross-validation of $M_0$ and then shared with $M_1$ to isolate the added feature rather than an added hyperparameter search. Coefficients in Table~\ref{tab:featurediagnostics} are on standardized development features. Their magnitudes are descriptive in the presence of correlated predictors; they should not be interpreted causally. An exploratory five-fold whole-corpus search for $M_1$ selects $\lambda=1$ and yields a held-out gain of $+.046$ nats ($[-.102,.171]$), so separate regularization does not materially change the conclusion.

\begin{table}[ht]
\caption{Measurements entering $M_0$ and $M_1$. Marginal correctness and $A_d$ reuse the same 48 demonstrator responses but summarize them differently.}
\label{tab:features}
\centering
\scriptsize
\setlength{\tabcolsep}{2.8pt}
\begin{tabular}{p{.14\linewidth}p{.37\linewidth}p{.14\linewidth}p{.17\linewidth}r}
\toprule
Feature & Definition & Same-state pairs? & Correctness labels? & Range \\
\midrule
Per-goal correctness & Fraction of goal-conditioned queries with the correct demonstrator branch & No & Yes & .438--.792 \\
Instruction entropy & Entropy of training instruction labels & No & No & 1.564--1.585 \\
Action entropy & Entropy of coarse training actions & No & No & .902--1.524 \\
$A_d$ & Fraction of paired states whose first choice changes & Yes & No & 0--.750 \\
\bottomrule
\end{tabular}
\end{table}

\begin{table}[ht]
\caption{Baseline feature ranges, correlation with $A_d$ over 136 corpora, and standardized coefficients. Checkpoint maturity is included only in the exploratory sensitivity.}
\label{tab:featurediagnostics}
\centering
\small
\setlength{\tabcolsep}{3.4pt}
\begin{tabular}{lrrrrr}
\toprule
Feature & Range & Pearson $r$ & Spearman $\rho$ & $M_0$ coef. & $M_1$ coef. \\
\midrule
Per-goal correctness & .438--.792 & .718 & .462 & .842 & .541 \\
Instruction entropy & 1.564--1.585 & $-.470$ & $-.454$ & $-.001$ & .113 \\
Action entropy & .902--1.524 & .815 & .564 & .966 & .549 \\
$A_d$ & 0--.750 & 1.000 & 1.000 & --- & 1.089 \\
$\log_{10}$ checkpoint episodes & 1.000--4.699 & .570 & .566 & --- & --- \\
\bottomrule
\end{tabular}
\end{table}

The primary baseline excludes checkpoint episode count because it is a simulator-specific experimental assignment rather than a measurement available for an arbitrary collected corpus. In an exploratory sensitivity, adding it to both nested predictors changes the test-corpus log-score gain to $+0.030$ nats (95\% corpus-bootstrap CI $[-0.036,0.090]$). This sensitivity does not strengthen the incremental case.

For each test corpus, the recorded 24 binary switch indicators are sampled with replacement 256 times. Each draw supplies one audit value to $M_1$; the 256 resulting beta--binomial mass functions are averaged before scoring. Development audit values enter this primary fit as observed features. The primary log score therefore integrates test-corpus finite-audit variability but does not propagate development-audit error through predictor fitting. The nested sensitivity in Appendix~\ref{app:robustness} performs that additional resampling and refit. The outer 10,000-draw bootstrap resamples whole corpora and serves a different purpose: uncertainty over the population of corpora.

\section{Prospective Corpus-Population and Precision Design}
\label{app:design}

The 56-development/80-test allocation, ordered corpus IDs, fitted predictors, and predictions were stored in a content-addressed internal commit before held-out outcomes were revealed. The accompanying certificate records the commit hash and forbids subsequent refitting. This is an auditable prospective record, not a public preregistration; the anonymized commit and certificate are included with the artifacts. The top-20\% operating point was prespecified, whereas its confidence interval was computed exploratorily.

Candidate allocations varied development corpora, test corpora, and learner seeds under a generative model fitted to preliminary development results. The effect coefficient was the smallest value on the prespecified grid whose calibration mean exceeded the 0.105-nat magnitude benchmark; the discrete grid produced a simulated mean gain of 0.174 nats. Among the three highest-power cells, the rule required estimated power of at least .82 and zero failed fits, then selected minimum Qwen compute, more test corpora, and higher power in that order. The chosen 56/80/two-seed cell had screening power .898 over 2,000 simulations. An independent 5,000-simulation confirmation gave .8934 (Monte Carlo 95\% interval $[.8848,.9020]$), with zero failed fits. This documents the design rule, not 89\% power for an effect of exactly 0.105 nats; only 12.5\% of the realized corpus-bootstrap replicates exceed that magnitude reference.

\begin{table}[ht]
\caption{Target corpus population. All varying seeds are disjoint across development and test sets; nested states and goals do not increase the independent-corpus count.}
\label{tab:corpuspopulation}
\centering
\small
\setlength{\tabcolsep}{4pt}
\begin{tabular}{p{.25\linewidth}p{.42\linewidth}p{.22\linewidth}}
\toprule
Factor & Levels or sampling & Status \\
\midrule
Corpus population & independently trained RL demonstrators & varying \\
Split & 56 development / 80 test & disjoint \\
Demonstrator checkpoint & 10, 50, 250, 1,000, 50,000 episodes & test: 16 each; dev.: 12/11/11/11/11 \\
Demonstrator and scene seeds & unique training, layout, physics, rendering seeds & varying \\
Goal families & three semantic object pairs, balanced sides & fixed design \\
Audit states & 24 paired states / 48 counterfactual queries & fixed quota \\
Training data & 48 unpaired multi-step trajectories & fixed quota \\
Evaluation data & 48 paired states per setting & fixed quota \\
Learner seeds & two shared seeds per corpus & crossed \\
\bottomrule
\end{tabular}
\end{table}

\section{Robustness of the Primary Conclusion}
\label{app:robustness}

For corpus $d$, estimate finite-state sampling variance as $A_d(1-A_d)/(n_A-1)$. If $s_A^2$ is the observed between-corpus variance, the classical variance-decomposition reliability is
\begin{equation}
  R_A=\max\!\left(0,1-
  \frac{|\mathcal D|^{-1}\sum_{d\in\mathcal D}A_d(1-A_d)/(n_A-1)}{s_A^2}
  \right).
  \label{eq:reliability}
\end{equation}
$R_A=1$ means estimated state-sampling noise explains none of the observed between-corpus variance; $R_A=0$ means it explains all of it. It is not a calibration curve or intraclass coefficient.

This estimate treats the 24 audit states within a corpus as exchangeable draws. It therefore does not capture additional dependence among states or uncertainty from unmodeled state strata.

Table~\ref{tab:robustness} separates the prespecified estimate from exploratory sensitivity checks. Reliability intervals resample corpus roots, and seed rows retain one crossed learner seed at a time. For the nested sensitivity, each replicate samples the 24 observed switch indicators with replacement within every development and test corpus, recomputes $A_d$, refits the unchanged $M_1$ specification on development outcomes, and evaluates the same test outcomes while holding $M_0$ fixed. A whole-corpus resample at evaluation propagates test-population uncertainty.

\begin{table}[ht]
\caption{Sensitivity of the primary Qwen2.5-ID conclusion. The prespecified row remains primary; all other rows are exploratory checks. Intervals resample whole corpora unless explicitly described as audit-measurement resampling.}
\label{tab:robustness}
\centering
\scriptsize
\setlength{\tabcolsep}{2.6pt}
\begin{tabular}{p{.39\linewidth}p{.27\linewidth}p{.27\linewidth}}
\toprule
Check & Result & Interpretation \\
\midrule
Audit reliability, all 136 & $.802\ [.688,.859]$ & stable corpus property \\
Audit reliability, test 80 & $.766\ [.523,.853]$ & stability persists on test corpora \\
Primary gain, pooled seeds & $+.059\ [-.027,.133]$ & positive, uncertain \\
Bootstrap replicates $>0$ & 92.3\% & positive direction \\
Bootstrap replicates $>.105$ & 12.5\% & 12.5\% exceed benchmark \\
Learner seed 20262001 & $+.067\ [-.051,.164]$ & positive, uncertain \\
Learner seed 20262002 & $+.052\ [-.025,.127]$ & positive, uncertain \\
Audit resampling + refit & $+.025\ [-.084,.111]$ & interval spans zero \\
Selected correctness, $M_1-M_0$ & $-.0266\ [-.0990,+.0234]$ & interval spans zero \\
Resampled selection, $M_1-M_0$ & $+.0019\ [-.0605,+.0430]$ & centered near zero \\
\bottomrule
\end{tabular}
\end{table}

The 92.3\% and 12.5\% values are proportions of bootstrap replicates, not posterior probabilities. The $-.0266$ primary selection difference evaluates the single prespecified pair of selected sets on observed outcomes. The $+.0019$ sensitivity repeatedly resamples audit states, refits $M_1$, and reselects corpora; it is therefore a different, exploratory measurement-error estimand. Its sign change reinforces that the selection effect is centered near zero rather than reversing the primary result. The audit coefficient remains positive across perturbations, but propagated log-score and selection effects remain uncertain.

\section{Exploratory Structural and Decision Diagnostics}
\label{app:officialdiagnostics}

These exploratory analyses examine identifiability, ranking resolution, and practical alternatives. They were conducted after held-out outcome release and do not modify the primary endpoint or prespecified predictions.

\paragraph{Audit decomposition:}
The audit responses are \textsc{left-30}, \textsc{right-30}, or \textsc{halt}. \textsc{Halt} constitutes 10.9\% of queries, and 107/136 corpora contain a nonbinary pair. Consequently, $p_d=.5+(a_d-b_d)/2$ holds exactly, while the binary-response identity $A_d=a_d+b_d$ does not: $A_d-(a_d+b_d)$ averages $.100$ and has maximum absolute value $.250$. Table~\ref{tab:decomposition} reports the two least mature checkpoints.

\begin{table}[ht]
\caption{Joint audit correctness at the two least mature checkpoints. $a$, $b$, and $c$ denote both correct, both wrong, and exactly one correct.}
\label{tab:decomposition}
\centering
\small
\begin{tabular}{rrrrrr}
\toprule
Episodes & Corpora & $a$ & $b$ & $c$ & Halt/query \\
\midrule
10 & 28 & .156 & .016 & .827 & .015 \\
50 & 27 & .389 & .074 & .537 & .098 \\
\bottomrule
\end{tabular}
\end{table}

\paragraph{Alternative features and selectors:}
Table~\ref{tab:posthocselectors} uses the same Qwen2.5-ID outcomes. The $a_d$ and $b_d$ rows add exact audit-derived values to $M_0$ at the common $\lambda=10$. Separate five-fold tuning selects $\lambda=1$ for $A_d$, $a_d$, and $b_d$, yielding $+.046$ ($[-.102,.171]$), $+.126$ ($[.037,.206]$), and $+.123$ ($[.026,.211]$) nats. This persistent ordering is post-hoc and does not isolate nonbinary responses from finite-sample or model-geometry effects. Training first-correct comes from accepted trajectories; acceptance is constant at 1.0 (6,528/6,528) and adds numerically zero. Qwen3.5 rows use collected learner outcomes rather than a refitted predictor.

\begin{table}[ht]
\caption{Exploratory Qwen2.5-ID diagnostics. Selected correctness, regret, and overlap use $K=16$; gain is relative to the prespecified $M_0$ where available.}
\label{tab:posthocselectors}
\centering
\scriptsize
\setlength{\tabcolsep}{3pt}
\begin{tabular}{lrrrrr}
\toprule
Method & Gain [95\% CI] & Spearman & Selected & Regret & Overlap \\
\midrule
$M_0+a_d$ & $+.104\ [.049,.153]$ & .560 & .846 & .070 & .312 \\
$M_0+b_d$ & $+.110\ [.022,.188]$ & .531 & .842 & .074 & .250 \\
$M_0+$ train first-correct & $+.156\ [.078,.225]$ & .584 & .831 & .085 & .312 \\
$M_0+$ acceptance & $+.000\ [.000,.000]$ & .589 & .838 & .078 & .312 \\
Demonstrator rollout success & --- & .281 & .838 & .078 & .312 \\
Qwen3.5 pooled outcomes & --- & .715 & .866 & .050 & .438 \\
Qwen3.5 seed 1 & --- & .644 & .846 & .070 & .500 \\
Qwen3.5 seed 2 & --- & .662 & .874 & .042 & .500 \\
\bottomrule
\end{tabular}
\end{table}

\paragraph{Selection resolution:}
Qwen2.5 learner-seed outcomes have Spearman $\rho=.567$. Selecting by seed~1 and evaluating on seed~2 gives $.874$ correctness, $.057$ regret, and $.50$ oracle overlap; reversing seeds gives $.874$, $.052$, and $.50$. After excluding 10- and 50-episode corpora, 48 roots remain. $M_0$ then has Spearman $.349$, selected correctness $.821$, and overlap $.20$; $M_1$ has $.172$, $.791$, and $.10$. Thus seed variation limits empirical resolution but does not account for the full $M_1$ ranking deficit.

\paragraph{Predictive calibration:}
Because $C_{d,s}$ is discrete, we report mid-PIT moments and central predictive-interval coverage over 160 corpus--seed rows. Both predictors are conservative (Table~\ref{tab:calibration}); a few severe log-score misses are not evidence of global undercoverage.

\begin{table}[ht]
\caption{Held-out Qwen2.5-ID predictive calibration.}
\label{tab:calibration}
\centering
\small
\begin{tabular}{lrrrrr}
\toprule
Model & Mid-PIT mean & Mid-PIT variance & 50\% cov. & 80\% cov. & 90\% cov. \\
\midrule
$M_0$ & .486 & .061 & .725 & .925 & .956 \\
$M_1$ & .487 & .054 & .763 & .938 & .963 \\
\bottomrule
\end{tabular}
\end{table}

\paragraph{Embodied target reach:}
Across the 16 rank-stratified closed-loop corpora, Spearman correlations with target reach are $.368$ for $A_d$, $.361$ for $M_0$, and $.423$ for $M_1$. These descriptive estimates use too few, deliberately stratified corpora to establish a target-reach selector.

\section{Complete Predictive and Selection Results}

\begin{figure}[ht]
  \centering
  \includegraphics[width=\linewidth]{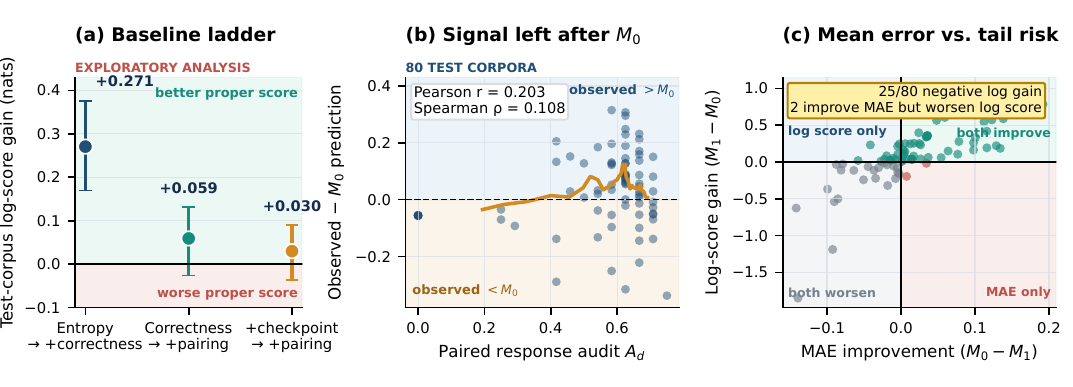}
  \caption{\textbf{Exploratory diagnostics of the remaining paired signal;} Light bands label the interpretation of each region. (a) Marginal per-goal correctness contributes substantially beyond corpus entropies, whereas the paired increment is smaller and uncertain. (b) $A_d$ has a weak relationship with the signed residual left by $M_0$. (c) The four regions distinguish improvements in MAE, proper log score, both, or neither; confident misses can lower MAE while worsening proper score.}
  \label{fig:incremental}
\end{figure}

\begin{table}[ht]
\caption{Test-corpus predictive results. Correct is mean paired correctness across test corpora and learner seeds. Gain is $M_1-M_0$ in nats; intervals resample whole corpora.}
\label{tab:predictiveappendix}
\centering
\small
\setlength{\tabcolsep}{4.2pt}
\begin{tabular}{llrrl}
\toprule
Backbone & Setting & Correct & Gain & 95\% CI \\
\midrule
Qwen2.5 & ID  & .708 & +.059 & $[-.027,.133]$ \\
Qwen2.5 & Shift & .801 & +.029 & $[-.030,.086]$ \\
Qwen3.5 & ID  & .761 & +.036 & $[-.005,.075]$ \\
Qwen3.5 & Shift & .776 & +.010 & $[-.010,.030]$ \\
\bottomrule
\end{tabular}
\end{table}

\begin{table}[ht]
\caption{Absolute prediction error and rank diagnostics. Lower absolute error is better; higher correlation is better.}
\label{tab:predictionappendix}
\centering
\small
\begin{tabular}{llrrrrrr}
\toprule
& & \multicolumn{2}{c}{Abs. error} & \multicolumn{2}{c}{Pearson} & \multicolumn{2}{c}{Spearman}\\
Backbone & Setting & $M_0$ & $M_1$ & $M_0$ & $M_1$ & $M_0$ & $M_1$\\
\midrule
Qwen2.5 & ID    & .110 & .096 & .887 & .900 & .589 & .490\\
Qwen2.5 & Shift & .106 & .097 & .908 & .921 & .460 & .391\\
Qwen3.5 & ID    & .077 & .073 & .927 & .934 & .752 & .745\\
Qwen3.5 & Shift & .080 & .078 & .923 & .925 & .729 & .723\\
\bottomrule
\end{tabular}
\end{table}

\begin{table}[ht]
\caption{Top-20\% corpus-selection diagnostics. Selected is realized mean correctness; regret is relative to the oracle top set. Regret and overlap are descriptive point estimates.}
\label{tab:triageappendix}
\centering
\small
\begin{tabular}{llrrrrrr}
\toprule
& & \multicolumn{2}{c}{Selected} & \multicolumn{2}{c}{Regret} & \multicolumn{2}{c}{Oracle overlap}\\
Backbone & Setting & $M_0$ & $M_1$ & $M_0$ & $M_1$ & $M_0$ & $M_1$\\
\midrule
Qwen2.5 & ID    & .838 & .811 & .078 & .105 & .312 & .250\\
Qwen2.5 & Shift & .945 & .917 & .055 & .083 & .250 & .188\\
Qwen3.5 & ID    & .965 & .971 & .031 & .026 & .562 & .625\\
Qwen3.5 & Shift & .969 & .964 & .031 & .036 & .438 & .438\\
\bottomrule
\end{tabular}
\end{table}

\begin{figure}[ht]
  \centering
  \includegraphics[width=\linewidth]{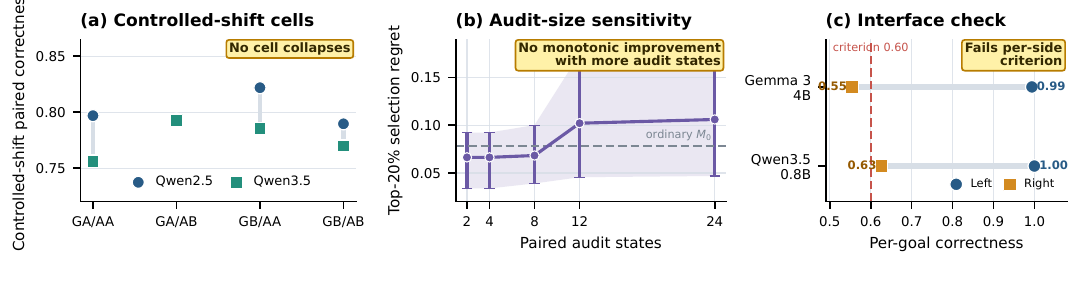}
  \caption{\textbf{Supporting analyses;} Controlled-shift performance remains substantial in every geometry/appearance cell; audit-size prefixes do not improve selection monotonically; and only models with an adequate instruction-conditioned action interface enter the test-corpus comparison.}
  \label{fig:supporting}
\end{figure}

\subsection{Development interface checks}
\label{app:interfacechecks}

Figure~\ref{fig:supporting}(c) shows the side-specific interface checks available for Gemma~3~4B and Qwen3.5. Gemma attains .99 left-goal but .55 right-goal correctness and fails the .60 per-side criterion; Qwen3.5 attains 1.00/.63 and advances to the corpus study. SmolVLM2 was evaluated in an earlier development screen rather than this side-specific check: 18 adapters achieve 31/432 paired decisions ($.0718$), compared with Qwen2.5's .6933 in the same screen. It was therefore not advanced to the corpus-level evaluation. These development diagnostics choose viable interfaces; they are not evidence for the primary predictive claim.

\begin{table}[ht]
\caption{Paired correctness in the balanced geometry $\times$ appearance shift.}
\label{tab:oodappendix}
\centering
\small
\begin{tabular}{lrr}
\toprule
Cell & Qwen2.5 & Qwen3.5\\
\midrule
Geometry A / appearance A & .797 & .756\\
Geometry A / appearance B & .794 & .793\\
Geometry B / appearance A & .822 & .785\\
Geometry B / appearance B & .790 & .770\\
\bottomrule
\end{tabular}
\end{table}

\section{Embodied and Physical Results}

\begin{figure}[t]
  \centering
  \makebox[\linewidth][c]{\includegraphics[width=\linewidth]{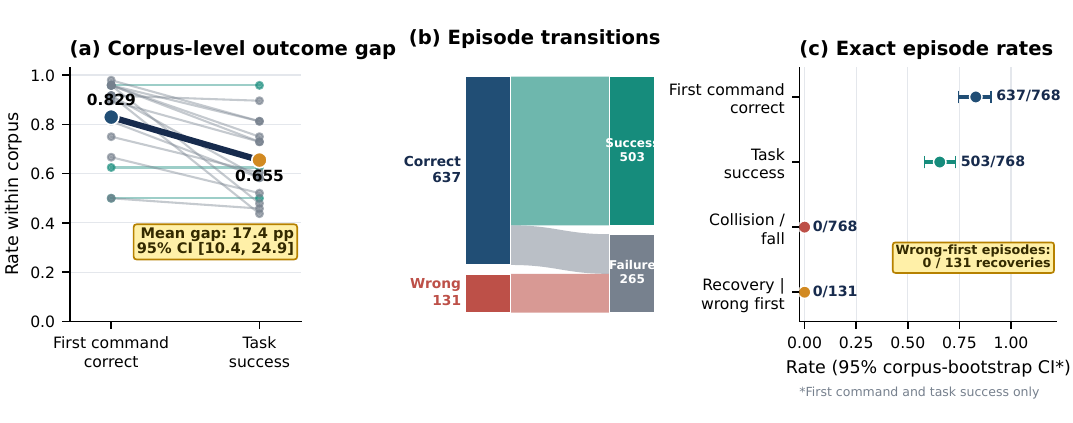}}
  \caption{\textbf{Local response and embodied completion diverge;} (a) Lines pair first-command correctness and target success within each of 16 rank-stratified corpora. (b) Exact episode flows expose 134 later failures after a correct first command. (c) First-command and task-success intervals resample corpus groups; zero collision/fall and recovery counts are descriptive.}
  \label{fig:embodied}
\end{figure}

\begin{figure}[t]
  \centering
  \includegraphics[width=\linewidth]{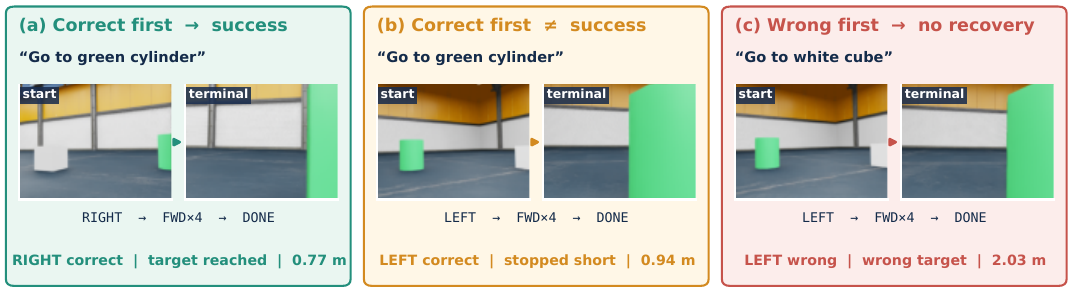}
  \caption{\textbf{A correct local decision can still fail downstream;} Representative simulated trajectories show success after a correct first command, failure after a correct first command, and failure after a wrong first command.}
  \label{fig:cases}
\end{figure}

Figure~\ref{fig:embodied} decomposes closed-loop outcomes at corpus and episode levels, while Figure~\ref{fig:cases} shows representative trajectories for the resulting success and failure modes.

The closed-loop decomposition is: 637/768 correct first commands, 503/768 target-reaching successes, 134 failures after a correct first command, and 131 failures after a wrong first command. A paired corpus bootstrap gives a 95\% interval of $[10.4,24.9]$ percentage points for the first-command minus task-success gap. Recovery is 0/131 and collision/fall is 0/768, with both zeros observed in every one of the 16 corpus groups; we report these as descriptive counts without episode-level binomial intervals. Mean path efficiency across the 16 corpora is .995 under the recorded metric, so it does not localize the 134 downstream failures.

\subsection{Physical closed-loop deployment}
\label{app:physical}

\begin{figure}[t]
  \centering
  \includegraphics[width=\linewidth]{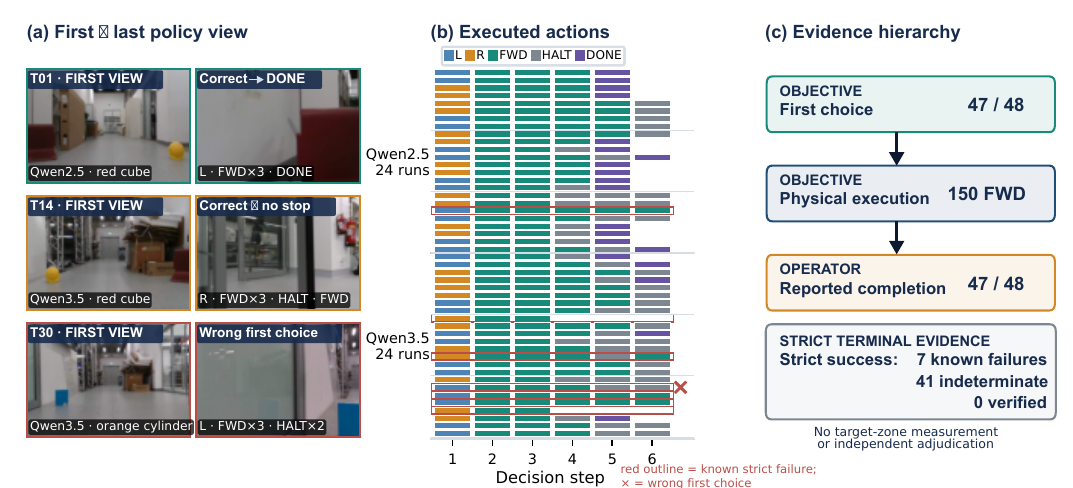}
  \caption{\textbf{Physical deployment transfers the local decision but does not establish strict terminal success;} (a) Representative first/last policy views. (b) Executed actions; red outlines mark seven known strict failures. (c) Recorder output, operator judgment, and terminal verification distinguish three evidential levels.}
  \label{fig:physical}
\end{figure}

Figure~\ref{fig:physical} summarizes the hierarchy from objective recorder outputs to operator judgment and unavailable terminal verification.

Table~\ref{tab:physical} separates objective recorder outputs, operator judgment, and strict terminal evidence. The first two rows share the same 47/48 count but not the same evidential status: first-choice correctness is reconstructed from the expected and executed branch commands, whereas completion is the contemporaneous operator report. No run has the external target-zone measurement and independent adjudication required to verify strict success. The resulting 0/48 verified count denotes missing verification, not 48 observed failures.

The 47/48 hardware first-choice rate should not be compared directly with the 82.9\% simulation-subset mean. Simulation averages 16 rank-stratified corpora, whereas hardware uses one median-selected corpus per backbone on six different physical configurations.

\begin{table}[ht]
\caption{Physical Go2 results for one fixed adapter per backbone. All counts are descriptive; the study was not designed to compare backbones.}
\label{tab:physical}
\centering
\small
\setlength{\tabcolsep}{4.1pt}
\begin{tabular}{lrrl}
\toprule
Endpoint & Qwen2.5 & Qwen3.5 & Evidence \\
\midrule
Correct first choice & 24/24 & 23/24 & recorder log \\
Operator-reported completion & 24/24 & 23/24 & operator judgment \\
Strict success & \multicolumn{2}{c}{not verifiable} & terminal evidence unavailable \\
Known strict failure & 1/24 & 6/24 & objective violation/report \\
Executed forward actions & 72 & 78 & controller log \\
Safety FWD$\rightarrow$HALT vetoes & 20 & 29 & depth-mask log \\
Mean summed forward motion & 1.54\,m & 1.63\,m & odometry \\
Communication aborts & 0 & 2 & recorder event \\
\bottomrule
\end{tabular}
\end{table}

The seven known strict failures comprise four runs that reach the six-decision horizon without a final \textsc{halt} or \textsc{done}, two communication/system aborts after motion, and the single run with a wrong first branch. The other 41 are indeterminate under the strict rule. For both backbones, the deployed corpus is the joint median of simulation-ID paired correctness over the 16 roots committed for closed-loop evaluation; the learner seed is fixed, and corpus-ID order resolves ties. No physical-image or physical-navigation outcome entered this choice; the exact corpus and seed remain in the reproducibility manifest. The selected Qwen2.5 adapter makes 12/12 correct first choices on each side; the selected Qwen3.5 adapter makes 12/12 on left goals and 11/12 on right goals. These nearly symmetric selected-policy counts do not contradict the aggregate stationary real-image result below, which averages 16 heterogeneous adapters.

\begin{figure}[t]
  \centering
  \includegraphics[width=\linewidth]{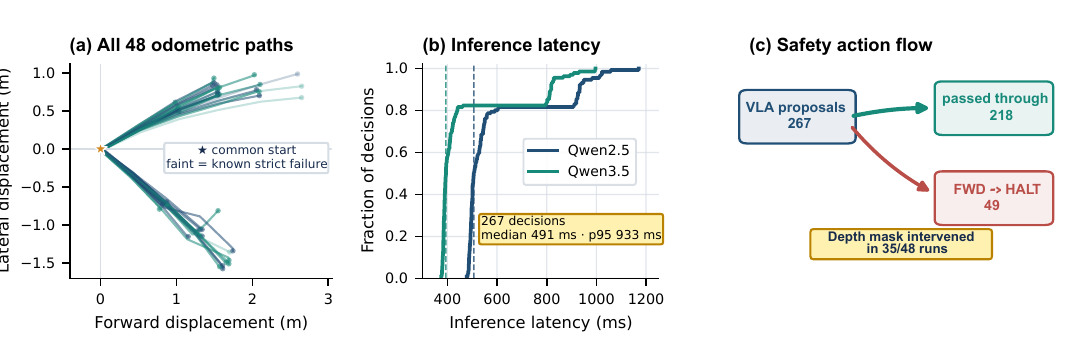}
  \caption{\textbf{The physical logs expose motion, latency, and safety mediation;} (a) All odometric paths after translating and rotating their starting poses to a common frame. (b) Empirical inference-latency distributions over 267 decisions; dashed lines mark model medians. (c) The depth mask changes 49 proposed forward actions to \textsc{halt}, affecting 35/48 runs. The study therefore evaluates a VLA--controller--safety stack rather than unconstrained VLA stopping.}
  \label{fig:physicaldiagnostics}
\end{figure}

Figure~\ref{fig:physicaldiagnostics} reports the recorded motion, inference latency, and depth-safety mediation for the deployed policy stack.

\begin{table}[ht]
\caption{Departures from the planned physical evaluation protocol. They restrict the deployment study to descriptive embodied-transfer evidence.}
\label{tab:physicaldeviations}
\centering
\small
\setlength{\tabcolsep}{3.5pt}
\begin{tabular}{p{.23\linewidth}p{.27\linewidth}p{.39\linewidth}}
\toprule
Element & Recorded study & Consequence \\
\midrule
Decision horizon & six rather than eight & terminal rule differs from plan \\
Trial status & 40 pilot; 8 unspecified & all 48 marked nonconforming \\
Instruction string & 17/48 non-exact & wording is not fully standardized \\
Terminal evidence & no external video, target coordinates, distance, or adjudication & strict success is not estimable \\
Configuration metadata & required pretrial files absent & geometric conditions cannot be fully reconstructed \\
Run order & grouped rather than counterbalanced & order effects are uncontrolled \\
Visual review & person in one start frame; tracked robot or duplicate prop in five runs & deployment scenes are not uniformly clean \\
System availability & two response-timeout aborts & two known strict failures \\
\bottomrule
\end{tabular}
\end{table}

\subsection{Physical-Go2 image strata}
\label{app:bankr}

Across the 36 stationary physical states, the same rank-stratified 16-adapter subset gives Qwen2.5 per-goal correctness .743, paired correctness .517, and instruction-switch rate .545; Qwen3.5 gives .811, .623, and .634. Exact parse rate is 1.0 for both. The left/right gaps (.958/.528 and .993/.628) motivate paired reporting: high aggregate per-goal correctness can conceal a side prior across adapters. Before inference, human review required a readable camera frame, both intended surrogate targets identifiable on the recorded sides, sufficient visibility to assign the first branch, and a valid identical-image instruction pair. All 36 prescreened images were retained, including documented surrogate-prop and incidental-scene variation; no model output entered screening. The props are semantic surrogates rather than replicas of simulated objects, so this setting tests external-image robustness only.

For the same 16 corpus adapters and learner seed, matched ID simulation states yield left/right first-command accuracies of .973/.677 for Qwen2.5 and .999/.749 for Qwen3.5. The asymmetry therefore predates physical transfer, while the physical images amplify its right-goal deficit. This comparison is descriptive for the rank-stratified subset and does not estimate prevalence over all 80 test corpora.

\begin{figure}[ht]
  \centering
  \includegraphics[width=\linewidth]{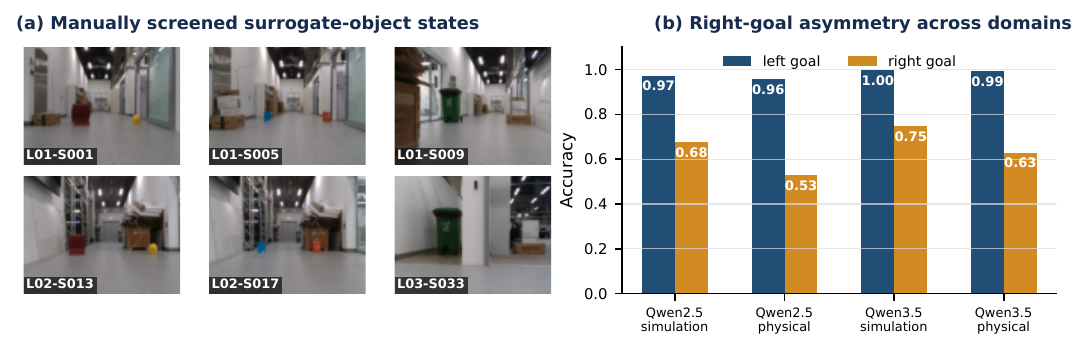}
  \caption{\textbf{Stationary physical images provide breadth across adapters;} (a) Six of 36 physical-Go2 states across three layouts and surrogate object pairs. (b) Matched first-command accuracy for the same 16 Qwen2.5 and Qwen3.5 corpus adapters in simulation and on physical images. The aggregate right-goal weakness does not imply that every selected adapter has the same asymmetry.}
  \label{fig:bankr}
\end{figure}

Figure~\ref{fig:bankr} summarizes stationary-image transfer across layouts, surrogate objects, and goal sides.

\section{Measured Cost}
\label{app:cost}

\begin{table}[ht]
\caption{Measured per-corpus cost. Rendering used one simulator GPU; adapter time is per fit.}
\label{tab:cost}
\centering
\small
\begin{tabular}{lrrl}
\toprule
Operation & Median & Mean & Output\\
\midrule
Paired audit rendering & 4.18 min & 4.78 min & 24 states / 48 queries\\
Training-trajectory rendering & 10.61 min & 11.35 min & 48 trajectories\\
Qwen2.5 adapter fit & --- & 10.68 min & one learner seed\\
Qwen2.5 evaluation generation & --- & 0.78 min & 48 goal-conditioned queries\\
Qwen3.5 adapter fit & 13.08 min & 13.41 min & one learner seed (121 measured fits)\\
\bottomrule
\end{tabular}
\end{table}

The full paired query set costs 48 demonstrator queries and a median 4.18 GPU-min per corpus. Once those responses have already been collected for the correctness-aware baseline, arranging them into same-state pairs and computing $A_d$ requires zero additional demonstrator queries and negligible arithmetic. Thus ``zero marginal query cost'' is conditional on existing correctness-labeled responses; a pipeline without them incurs the full acquisition cost above. In this small-data regime, one Qwen3.5 proxy fit is not cheaper than one Qwen2.5 target fit plus evaluation (13.41 versus 11.46 mean GPU-min). Costs exclude one-time model download, rendering, and environment setup.

\section{Reproducibility Details}
\label{app:reproducibility}

Within each backbone, the revision, prompt, processor, optimizer, update budget, decoder, grammar, and controller are fixed; Table~\ref{tab:learnerconfig} gives their exact values. We preserve all 136 corpus definitions, 544 adapter configurations, seeds, ordered partitions, asset hashes, 256 audit draws per corpus, the 10,000-draw primary bootstrap, $\lambda=10$, and the 0.105-nat benchmark. Audit, training, and evaluation observations are disjoint. Predictions and rules were prespecified before test outcomes, with no outcome-dependent exclusion. Higher-draw robustness artifacts and all exploratory or descriptive analyses are separately labeled and do not alter the primary test.

\begin{table}[ht]
\caption{Learner and demonstrator configuration held fixed within each backbone. Qwen3.5 is a supporting replication under the same learner recipe.}
\label{tab:learnerconfig}
\label{tab:teacherconfig}
\centering
\small
\renewcommand{\arraystretch}{1.08}
\setlength{\tabcolsep}{3pt}
\begin{tabular}{p{.25\linewidth}p{.67\linewidth}}
\toprule
Item & Configuration \\
\midrule
Base artifacts & Qwen2.5-VL-3B-Instruct, Qwen3.5-VL-0.8B. \\
Input processor & One current RGB frame per decision; Qwen pixel cap 102,400; architecture-native chat template; action history serialized oldest first. No explicit token truncation beyond the base processor/model limit. \\
Quantization & NF4 4-bit weights, double quantization, bfloat16 compute; vision tower held fixed. \\
LoRA & Rank 8, alpha 16, dropout .05, no bias; language modules \texttt{q/k/v/o\_proj}, \texttt{gate/up/down\_proj}; all visual modules excluded and asserted absent. \\
Optimization & AdamW (PyTorch), learning rate $2\times10^{-4}$, cosine schedule, zero warmup, weight decay 0, batch size 1, gradient accumulation 8, three epochs (105--135 optimizer updates as corpus steps vary from 273--358). \\
Decoding & Greedy (no sampling), at most 16 new tokens; strict five-class grammar \{forward, left, right, halt, done\}; invalid generations count as errors. \\
Seeds & Learner seeds \texttt{20262001} and \texttt{20262002} cross every corpus; Python, NumPy, and all CUDA generators receive the learner seed. \\
Demonstrator & Tabular Q-learning with deterministic argmax; state is RGB-derived target bearing $\{-1,0,+1\}$ and three binary depth-clearance sectors; actions are \textsc{left-30}, \textsc{right-30}, and \textsc{halt}; no target coordinates or branch labels are inputs. \\
Reward / maturity & Branch reward $2$ if desired and $-1$ otherwise, plus $0.5$ for a safe sector and $-2$ otherwise; checkpoints at 10, 50, 250, 1,000, and 50,000 episodes; held-out allocation 16 each and development allocation 12/11/11/11/11. \\
Corpus rollout & Unique teacher, layout, physics, rendering, neighborhood, and outcome seeds; at most eight decisions, 0.8\,m reach radius, termination on \textsc{done} or collision/fall, timeout otherwise. Structural validity requires nonempty fresh observations, consistent histories, permitted actions, and finite translations; correctness and success are not acceptance gates. All 6,528 attempts were accepted. \\
\bottomrule
\end{tabular}
\renewcommand{\arraystretch}{1.0}
\end{table}

The exact prompt is version \texttt{go2\_vla\_prompt/v1}: \emph{``You are the navigation policy of a Unitree Go2 quadruped robot. Instruction: `\{instruction\}'. The image is the current view from the robot's forward camera. Your recent actions, oldest first: \{history\}. Reply with exactly one next action and nothing else''}, followed by the five canonical forms for forward 25/50/75\,cm, left or right 15/30/45 degrees, hold position, and instruction completion. The same builder and parser are used for training and deployment. During corpus rollout, a valid initial turn is followed by \textsc{forward-50}, then \textsc{done} after the reach condition.